%% file: pprai.tex
\documentclass{pprai}

\usepackage[utf8]{inputenc}
\usepackage[T1]{fontenc}
\usepackage{graphicx}
\usepackage{color}
\usepackage{amsthm}
\usepackage{txfonts}
\usepackage{url}
\usepackage{algpseudocode}
\usepackage{listings}

\title{Model Discovery with Grammatical Evolution. An Experiment with Prime Numbers}
\headtitle{Model Discovery with Grammatical Evolution}


\author{Jakub Skrzyński$^{1[0009-0008-4550-5009]}$, Dominik Sepioło$^{1[0000-0001-7746-3781]}$, \\ Antoni Ligęza$^{1[0000-0002-6573-4246]}$}
\headauthor{J. Skrzyński, D. Sepioło, A. Ligęza}
\affiliation{%
 $^1$AGH University of Krakow\\
  Department of Applied Computer Science\\
  al. A. Mickiewicza 30, 30-059 Krakow, Poland\\
  jskrzynski@student.agh.edu.pl, \{sepiolo, ligeza\}@agh.edu.pl}


\begin{document}
\maketitle

\begin{abstract}
Machine Learning produces efficient decision and prediction models based on input-output data only. Such models have the form of decision trees or neural nets and are far from transparent analytical models, based on mathematical formulas. Analytical model discovery requires additional knowledge and may be performed with Grammatical Evolution. Such models are transparent, concise, and have readable components and structure. This paper reports on a non-trivial experiment with generating such models.
\end{abstract}

\section{Introduction}
The need for development of transparent, white-box, explainable models in Artificial Intelligence becomes more and more visible \cite{DBLP:journals/natmi/Rudin19}. In Machine Learning (ML), the classical decision trees induction and neural nets learning supplemented with the shallow methods of post-hoc eXplainable Artificial Intelligence (XAI) \cite{arrieta2019explainable} might be replaced by Model-Driven approaches \cite{Hydra23}, \cite{pprai23}. Model-Discovery seems to be a promising direction in Artificial Intelligence in domains where understanding \textit{how-it-works} is required \cite{DBLP:conf/ismis/LigezaJASAJKS020,AI4S:2023}.


One of the methods enabling development of transparent models is \textit{Grammatical Evolution} (GE) \cite{o2001grammatical, HandbookGE}. GE is a kind of genetic algorithm that can perform \textit{Symbolic Regression} tasks and find the best combination of operations to discover a formula that describes a given phenomenon as accurately as possible. 
 
 GE is an evolutionary algorithm that utilizes additional knowledge, provided as a Context-Free Grammar (CFG), to perform genotype-phenotype mapping, allowing limiting the search space and simultaneously increasing the search efficiency. CFG design has a direct impact on search efficiency and correctness of given solutions, but at the same time, it gives huge flexibility in terms of describing the problem. Depending on CFG, the result could be a mathematical expression, architecture of a neural network, or a syntactically correct computer program. GE is a candidate approach for achieving explainability better than some shallow approaches \cite{pprai23}.

The main aim of this paper is to demonstrate the use of GE in search of white-box model based on existing partial knowledge of problem structure, expressed as CFG, and relatively small sets of collected data. For the purpose of the experiment, the function known for lack of exact formula was chosen to emphasize the approximation capabilities of GE.

The approach outlined in this paper can be applied in real-life scenarios 
where estimation of an unknown function is needed with only partial knowledge available. This method is particularly useful for identifying relationships between variables in complex phenomena that are difficult to analyze analytically. Examples include the discovery of relation between medical parameters, or describing phenomena observed in physics and much more.

\section{Experiment}
Function $\pi(x)$ is defined as the amount of \textit{prime numbers} less or equal to $x$ \cite{shanks2001solved}. Despite its significance, the precise calculation of the $\pi(x)$ function remains a challenging endeavor due to the nature of prime numbers and their distribution. The elusive and irregular pattern of prime numbers presents a formidable computational hurdle, leading to the absence of an exact analytical expression for $\pi(x)$.

Grammatical Evolution emerges as a promising method for evolving mathematical formulas that endeavor to capture the intricate nature of prime number distribution. Unlike traditional approaches, Grammatical Evolution harnesses the power of evolutionary algorithms to iteratively refine and optimize mathematical expressions, seeking a formula that best approximates $\pi(x)$.

A series of experiments, conducted using the Python programming language leveraging the PonyGE2 library \cite{pony}, aimed to construct models that seek to unveil the $\pi(x)$ function, based on example data and mean squared error (MSE) as the fitness function, is reported.

\subsection{The Grammar}

To evolve formulae approximating $\pi(x)$, a grammar describing mathematical expressions is needed. 
For demonstration purposes, we utilized the grammar provided by PonyGE2 with slight modifications. These adjustments, made to align with the specific requirements of our experiment, are presented below:

\begin{lstlisting}
<e>  ::=  <e>+<e>|<e>-<e>|<e>*<e>|        
          pdiv(<e>,<e>)|
          psqrt(<e>)|
          np.sin(<e>)|
          np.tanh(<e>)|
          np.exp(<e>)|
          plog(<e>)|
          x[:, 0]|
          <c><c>.<c><c>
<c>  ::= 0 | 1 | 2 | 3 | 4 | 5 | 6 | 7 | 8 | 9
\end{lstlisting}

\subsection{The Data Set}
In order to generate an example dataset, a C++ program was implemented. The program produces a hard-coded array of prime numbers which can be referred to during operation to optimize time consumption. Prime numbers may be taken from internet sources. Generated data set quality depends on the number of prime numbers. Data generated by the program should be stored in a text file, in a format required by PonyGE2, that will be later referenced in the parameters file.  

To prepare the dataset, a subset of prime numbers was selected, including primes within the range of  $<2;7919>$ , resulting in a total of 1000 entries.

\section{The Results}

Throughout the experiment, multiple functions resembling the shape of $\pi(x)$ were evolved. However, a challenge arises in the complexity of the evolved formulas, characterized by numerous nested operations. An example of one such result is presented below:


\begin{lstlisting}
2 sqrt(x) + x/(tanh((x + sqrt(tanh(78.45) sin(51.98)) x - log(sqrt(84.76) + 47.5))/exp(log(log(69.92) + 7.51)) x) + sqrt(38.86) + log(log(x - log(sin(x) + 15.6) tanh(tanh(sqrt(x))) tanh(sin(log(x)) + 69.37) x)))
\end{lstlisting}

The complexity issue can be addressed, for example, by imposing stricter constraints on the depth of the derivation tree. However, this approach may affect the accuracy of the results.

\begin{figure}[htb]
\centering
\includegraphics[width=0.85\linewidth]{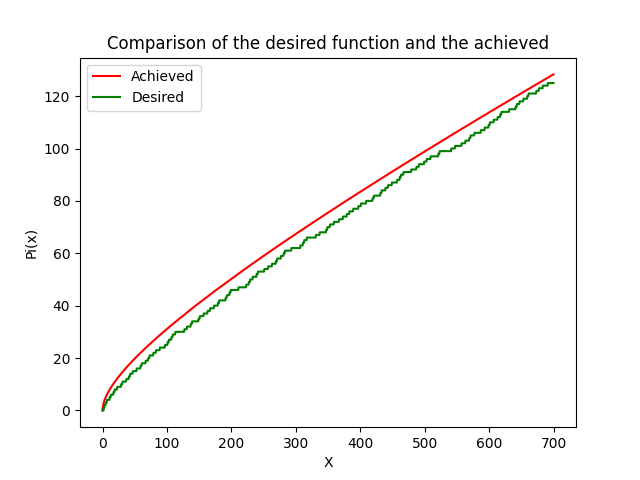}
\caption{First solution generated with GE compared to actual function.}\label{fig:pi_of_x_vs_solution}
\end{figure}

As it can be seen in Fig.~\ref{fig:pi_of_x_vs_solution}, the generated solution does have a similar shape to the desired function. However, the approximation is not perfect, showcasing differences between predicted values and actual ones. Subsequent executions of the experiment, with a higher number of generations, were conducted, yielding the following results:

\begin{lstlisting}
x/(ln(x/(ln(ln(92.89-sin(x)+x*x+sin(x)-64.03*sqrt(x)*ln(exp(sin(89.77))))*sqrt(sin(19.94))))))
\end{lstlisting}

\begin{figure}[htb]
\centering
\includegraphics[width=0.85\linewidth]{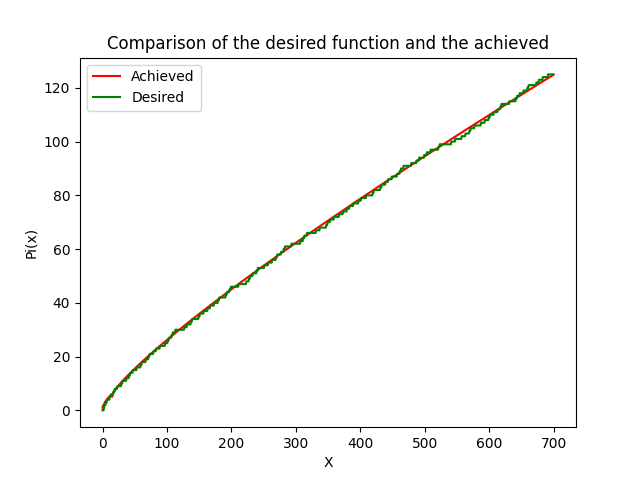}
\caption{Second generated solution in comparison to actual function.}\label{fig:pi_of_x_vs_solution_2}
\end{figure}

While the results of the simple experiment are reasonably satisfying, differences persist. Table \ref{tab:f2_vs_pi} compares function values for two $x$ values, highlighting that the evolved function is close but does not precisely describe the function. To gain a visual overview, one may refer to Fig.~\ref{fig:pi_of_x_vs_solution_2}.

\begin{table}[!ht]
    \centering
    \begin{tabular}{|c|c|c|}
    \hline
        $x$ & $f_2(x)$ & $\pi(x)$\\
        \hline
        \hline
         100 & 26.0574 & 25 \\
         1400 & 222.801 & 222 \\
         \hline
    \end{tabular}
    \caption{Values of $\pi(x)$ in comparison to values of second evolved function}
    \label{tab:f2_vs_pi}
\end{table}

The results of the experiment show greater promise in terms of execution time, with the evolution of the second formula taking only 143.7 seconds. Considering the achieved effectiveness relative to resource consumption, there is a potential to enhance precision by expanding the search space. This could lead to an even more refined approximation of the given function. Furthermore, supplying a larger dataset might help maintain the function shape across a broader range of values.

\section{Further work}

Considering the successful application of performing regression in order to find the approximate form of an unknown function of one parameter, further research may focus on applications of this technique to more complex scenarios. This includes examining relationships involving multiple independent variables and one or more dependent variables, with a constant focus on enhancing accuracy.

\section{Conclusions}

The performed experiments show just one of many potential use cases, but emphasize the advantages of GE and also potential issues that may emerge during the development of the GE model. Grammars (and their definitions) are candidate tool to introduce additional knowledge into the system. Careful design may limit the search space in accordance with already possessed knowledge about the system. 

Designing a correct model requires deep knowledge of the utility of the solution. It allows the user to create a correct fitness function that is being used to rate evolved solutions, therefore allowing the selection of the correct results.

GE models allow injecting knowledge into the system and have the efficiency of retrieving good solutions proportional to the amount of introduced knowledge. GE does not require a lot of theory knowledge from the final users, enabling them to adopt the technique relatively quickly.

\bibliography{pprai}
\bibliographystyle{pprai}

\end{document}